\documentclass[11pt, a4paper,notitlepage]{article}
\usepackage{amsfonts, amsmath, hyperref, parskip, hanging, times}%
\usepackage{graphicx}	
\usepackage{inputenc}
\usepackage{float}
\usepackage{verbatim} 
\usepackage{enumerate}
\usepackage{enumitem}   
\usepackage{multirow}
\usepackage{bigstrut}
\usepackage{listings}
\usepackage{array}

\usepackage[toc,page]{appendix}   
\usepackage{tikz}
\usetikzlibrary{arrows,chains,matrix,positioning,scopes}
\usepackage{pgfplots}
\usepackage{fancyvrb}  
\usepackage[skip=2pt]{caption}
\usepackage[skip=0pt]{subcaption}
\DeclareCaptionFont{tiny}{\tiny}
\usepackage{tikz}
\usetikzlibrary{calc,trees,positioning,arrows,chains,shapes.geometric,%
    decorations.pathreplacing,decorations.pathmorphing,shapes,%
    matrix,shapes.symbols}

\usepackage{soul} 

\renewcommand{\title}[1]{\begin{center}{\bf \LARGE #1}\end{center}}

\begin{document}

\pagestyle{empty}

\title{Deep learning for assessing banks' distress from news and numerical financial data}
{
\begin{center}

\vspace*{\stretch{0.8}} \par
\vspace*{\stretch{1.6}}

\Large
{\bf  P. Cerchiello$^\dag$,  G. Nicola$^{\dag 1}$ S. R\"onnqvist$^*$ and P. Sarlin$^{+ \ddag}$ } \\[2mm]
\normalsize
$^\dag$ Dep. of Economics and Management Science, University of Pavia.  \\
$^*$ Turku Centre for Computer Science -- TUCS, \r{A}bo Akademi University, Turku, Finland.\\
$^+$ Department of Economics, Hanken School of Economics, Helsinki, Finland.\\
$^\ddag$ RiskLab Finland, Arcada, Helsinki, Finland.

\par
\vspace*{\stretch{1.5}}
\begin{abstract}
 In this paper we focus our attention on the exploitation of the information contained in financial news to enhance the performance of a classifier of bank distress. Such information should be analyzed and inserted into the predictive model in the most efficient way and this task deals with the issues related to text analysis and specifically to the analysis of news media.

Among the different models proposed for such purpose, we investigate one of the possible deep learning approaches, based on a doc2vec representation of the textual data, a kind of neural network able to map the sequence of words contained within a text onto a reduced latent semantic space.
Afterwards, a second supervised neural network is trained combining news data with standard financial figures to classify banks whether in distressed or tranquil states.
Indeed, the final aim is not only the improvement of the predictive performance of the classifier but also to assess the importance of news data in the classification process. Does news data really bring more useful information not contained in standard financial variables? Our results seem to confirm such hypothesis.

\vskip 0.5cm
\vspace{0.3cm}

\vskip 0.5cm
\vspace{0.3cm}

\end{abstract}
\vspace*{\stretch{1.5}}

\end{center}
\vspace*{\stretch{0.5}}

\begin{affiliations}
$^1$ Corresponding Author\\
\href{mailto:giancarlo.nicola01@universitadipavia.it}{giancarlo.nicola01@universitadipavia.it}\\[-2pt]

\end{affiliations}

\newpage
\thispagestyle{empty}
\pagestyle{plain}
\setcounter{page}{2}
\pagenumbering{arabic}

\mbox{}
\

\section{Introduction}
\null\vspace{\stretch{1}}
\vspace{\stretch{2}}\null

Natural Language Processing (NLP), the interpretation of text by machines, is a complex task due to the richness of human language, its highly unstructured form and the ambiguity present at many levels, including the syntactic and semantic ones. From a computational point of view, processing language means dealing with sequential, highly variable and sparse symbolic data, with surface forms that cover the deeper structures of meaning.

Despite these difficulties, there are several methods available today that allow for the extraction of part of the information content present in texts. Some of these rely on hand crafted features, while others are highly data-driven and exploit statistical regularities in language. 
Moreover, once the textual information has been extracted, it is possible to enhance it with contextual information related to other sources different from text. The introduction of contextual information in the models is not always a straightforward process but requires a careful choice of the additional information provided in order to not increase noise by using irrelevant features. To accomplish such purpose, there are several methods of variable selection (Guyon and Elisseeff 2003) that can guide in the choice of the additional features for the model. The recent advancements in text analytics and the addition of contextual information aim at increasing the potential value of text as a source in data analysis with a special emphasis on financial applications (see for example Nyman et al. 2015). In this paper, we focus on the issues of understanding and predicting banks distress, a research area where text data hold promising potential due to the frequency and information richness of financial news. Indeed, central banks are starting to recognize the usefulness of textual data in financial risk analytics (Bholat et al. 2015, Hokkanen et al. 2015).

If we focus only on the elicitation of information from textual data, we can find that among the statistical methods, many rely on word representations. Class based models, for example, learn classes of similar words based on distributional information, like Brown clustering (Brown et al. 1992) and Exchange clustering (Martin et al. 1998, Clark 2003). Soft clustering methods, like Latent Semantic Analysis (LSA) (Landauer et al. 1998) and Latent Dirichlet Allocation (Blei et al. 2003), associate words to topics through a distribution over words of how likely each word is in each cluster. In the last years many contributions employ machine learning and semantic vector representations (Mikolov et al. 2013, Pennington et al. 2014), lately using Long Short-Term Memory (LSTM) networks (Hochreiter and Schmidhuber 1997, Socher et al. 2013, Cho et al. 2014) to model complex and non-local relationships in the sequential symbolic input. Recursive Neural Tensor Networks (RNTN) for semantic compositionality (Socher et al. 2011, Socher et al. 2013) and also convolutional networks (CNN) for both sentiment analysis (Collobert et al. 2011) and sentence modelling (Kalchbrenner 2014). In this vein, (Mikolov et al. 2013), (Mikolov 2012) and (Pennington et al. 2014) have introduced unsupervised learning methods to create a dense multidimensional space where words are represented by vectors. The position of such vectors is related to their semantic meaning, further developing the work on word embeddings (Bengio et al. 2003) which grounds on the idea of distributed representations for symbols (Hinton et al. 1986). The word embeddings are widely used in modern NLP since they allow for a dimensionality reduction compared to a traditional sparse vector space model. In (Le and Mikolov 2014), expanding the previous work on word embeddings, is presented a model capable of representing also sentences in a dense multidimensional space. Also in this case sentences are represented by vectors whose position is related to the semantic content of the sentence. In such a space sentences with similar semantic will be represented by vectors that are close to each other.
\\
\\

This recent rise of interest around text-based computational methods for measuring financial risk and distress is fuelling a rapidly growing literature (e.g. Cerchiello et al. 2017). The most covered area is sentiment analysis to be correlated with events of interest. Many of the previous approaches have been based on hand-crafted dictionaries that despite requiring work to be adapted to single tasks can guarantee good results due to the direct link to human emotions and the capability of generalizing well through different datasets. Examples of this kind are the papers of Nyman et al. 2015 and Soo 2013. The first analyses sentiment trends in news narratives in terms of excitement/anxiety and find increased consensus to reflect pre-crisis market exuberance, while the second correlates the sentiment extracted from news with the housing market. Despite the good results, there are applications where it could be preferable to avoid dictionaries in favour of more data driven methods, which have the advantage of higher data coverage and capability of going beyond single word sentiment expression. Malo et al. 2014 provide an example of a more sophisticated supervised corpus-based approach, in which they apply a framework modelling financial sentiment expressions by a custom data set of annotated phrases. 
\\
\\
Our contribution aims at demonstrating the feasibility and usefulness of the integration of textual and numerical data in a machine learning framework for financial predictions. Thus, the goal of the predictive model is to correctly classify stressed banks from both financial news and financial numerical data. 

The paper is organized as follows: in Section 2 we describe the machine learning framework, in Section 3 we illustrate the data and the predictive task, in Section 4 we present the experimental results with a sensitivity analysis on the network parameters and in Section 5 we show the conclusions of the work with hints on future developments.

\section{Deep learning framework}

Machine learning systems benefit from their ability to learn abstract representations of data, inferring feature representations directly from data instead of relying on manual feature engineering. This capability is particularly exploited in deep learning models, which provides flexibility and potentially better accuracy (Schmidhuber 2015). In particular the flexibility is crucial in natural language processing tasks where the ability to generalize across languages, domains and tasks enhances the applicability and robustness of text analysis.
The framework applied in this paper is an extension of the one developed in (R\"onnqvist and Sarlin 2017) with the aim of predicting banks' distress from textual data. The approach allows inferring banks distress conditions from textual news with a machine learning approach based on two steps:
\begin{itemize}
\item The first step comprises an unsupervised algorithm to retrieve the semantic vectors associated to a text. Using the Distributed Memory Model of Paragraph Vectors (PV-DM) by (Le and Mikolov 2014) (here referred to as doc2vec) for learning vector representations of sequences of words, dense vector representations of sentences mentioning target banks are learned. This method creates a multidimensional space (600 dimensions), in which words are positioned according to their semantic meaning (the closer the meaning of two words, the closer are their positions). From this new space is easier to perform the classification task due to the reduced dimensionality and the wise positioning of the vectors that takes into account their semantic meaning.
\item The second step performs the classification task through a supervised algorithm. The sentence vectors are fed into a neural network classifier. The neural network is composed of one input layer (600 nodes), one hidden layer (50 nodes) and one output layer (2 nodes with stress prediction $e \in \lbrace0, 1\rbrace$).
\end{itemize}

In this paper we modify the previous model of (R\"onnqvist and Sarlin 2017) to integrate the support of financial numerical data. The financial data that we integrate hold information about bank accounting data, banking sector data and country macroeconomic data. The two-step structure of the data processing flow has been preserved. 

The approach used to learn the semantic vectors is a Distributed Memory Model Paragraph Vector (Le and Mikolov 2014). In this model the semantic vector representation is learned by training a feed forward neural network to predict a certain word using its word context (previous $n$ and following $n$ words) and a randomly initialized semantic vector (sentence $ID$). The word contexts, used as features to predict the target words are fixed-length and sampled from a sliding window over the sentence. While training the network, the semantic vector gets updated by the training algorithm so that its representation positively contributes in predicting the next word and thus works as a semantic representation of the entire sentence. In this way the sentence $ID$ works as a memory for the model that allows the vector to capture the semantics of continuous sequences. The sentence $ID$, in fact, can be thought of as an extra word representing the sentence as global context and conditioning the prediction of the next word. Despite the random initialisation of semantic vectors, they gradually improve the capability of capturing the semantic of the sentence during the training, performed by stochastic gradient descent and with the gradient computed via backpropagation algorithm. Formally, the training procedure seeks to maximize the average log probability:

\begin{center}
\begin{equation}
\dfrac{1}{t+n}\sum_{i=1}^{t-n} \log \textit{p}\!\left( w_{i+n+1}|s, w_{i},...,w_{i+n}\right) 
\end{equation}
\end{center}

over the sequence of training words $w_{1}, w_{2},..., w_{t}$  in sentence $s$ with word context of size $n$.
The dimensionality of the semantic vectors and the context size of the algorithm have been optimized by cross-validation.
\\
\\
The second step, performing the classification task, receives in input two different sources: news textual data on the banks, in form of multidimensional semantic vectors $V_{s}$, and numerical financial data $F_{s}$ loaded from a database matched with news data. For the classification task we employ a three layers fully connected feed forward neural network. The neural network has a 612 dimensional input layer - 600 input nodes for the semantic vector $V_{s}$ dimensionality and 12 input nodes for the numerical data $F_{s}$ dimensionality - 50 hidden nodes and 2 output nodes for $e \in \lbrace0, 1\rbrace$ encoding the distress or tranquil status in a softmax layer that applies a cross-entropy loss function (see fig. 1). Finally, an additional phase of cleaning and merging of textual and numerical data has been added. The network is trained by Nesterov’s Accelerated Gradient Descent (Nesterov 1983) to predict distress events $e \in \lbrace0, 1\rbrace$. Hence, the objective is to maximize the average log probability:

\begin{center}
\begin{equation}
\dfrac{1}{\vert S \vert}\sum_{s \in S} \log \textit{p}\!\left( e_{s}|V_{s},F_{s}\right)
\end{equation}
\end{center}

In the trained network, the posterior probability $\textit{p} \left(e_{s} = 1| V_{s},F_{s}\right)$ reflects the relevance of sentence s to the modelled event type.

\begin{figure}[h!]
\centering
\includegraphics[scale=0.4]{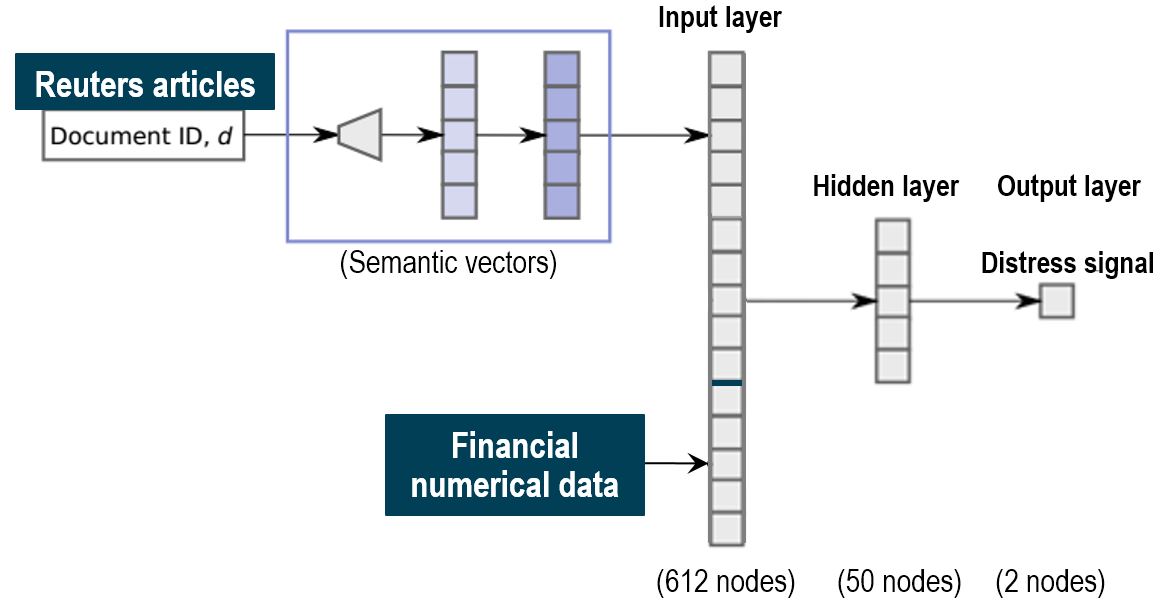} 
\caption{Structure of the model}
\end{figure}

\section{The Data}

As described in the previous section, we leverage two types of data, textual and numerical, matched by time and entities (banks), to predict events of interest. The events dataset contains information on dates and names of involved entities, relating to the specific type of event to be modelled, while the textual and financial numerical datasets contain respectively bank related articles and financial figures. The textual and numerical data are connected to a particular event using date and occurrences of the entity name for the textual data and matching the date with the corresponding quarter for the financial numerical data. The model is then trained in a supervised framework to associate specific language and financial figures with the target event type.

\subsection{Textual and numerical data }

The textual data are taken from a database of news articles from Reuters on-line archive spanning the years from 2007-Q1 to 2014-Q3. The original data set includes 6.6M articles, for a total of ca. 3.4B words. In order to select only articles related to the banks object of study, we have looked at bank name occurrences and selected only those articles with at least one occurrence. Bank name occurrences are located using a set of patterns defined as regular expressions that cover common spelling variations and abbreviations of the bank names. The regular expressions have been iteratively developed on the data to increase accuracy, with a particular attention on avoiding false positives. As a result from the entire corpus we retrieve 262k articles mentioning any of the 101 target banks. Successively the articles are split into sentences and only the sentences with bank name occurrences are kept. 
We integrate contextual information through a database of distress related indicators for banks. The numerical dataset is composed of 12 variables for 101 banks over the period 2007-2014 with quarterly frequency. In table 1 we report the list of numerical variables: among them there are information on bank-level balance sheet and income statement data, as well as country-level banking sector and macro-financial data.

The three bank-specific variables are the ratio of tangible equity to total assets, the ratio of interest expenses to total liabilities and the NPL reserves
to total assets ratio. The three banking-sector features are the mortgages to loans ratio (4-months change), the ratio of issued debt securities to total liabilities (4-months change) and the ratio of financial assets to GDP. The six Macro financial level features are the House price gap (Deviation from trend of the real residential property price index filtered with the Hodrick-Prescott Filter (Hodrick and Prescott 1980) with a smoothing parameter $\lambda$ of 1600), the  international investment position from the ECB Macroeconomic Imbalance Procedure (MIP) Scoreboard, the country private debt, the government bond yield, the credit to GDP ratio and the credit to GDP 1-yr change.

\begin{table}[!htbp] \centering 
	\label{} 
	\tiny
	\resizebox{\columnwidth}{!}{%
		\begin{tabular}{ l l l } \\ [-2.8ex]
			\textbf{Bank Level} & \textbf{Bank Sector Level} & \textbf{Macro Level} \\ \hline \\[-1.8ex] 
			Capital to asset & Mortgages to loans & House price gap (Deviation from trend of the real residential property price index)  \\ 
			Interest to liabilities & Securities to liabilities d4 &  Macroeconomic Imbalance Procedure (MIP), international investment position \\ 
			Reserves to asset & Financial assets to gdp & Private debt \\ 
			- & - & Government bond yield \\ 
			- &  - & Credit to gdp \\ 
			- &  - & Credit to gdp delta over 12 months \\ \hline 
		\end{tabular} 
	}
	\caption{List of available numerical variables. } 
\end{table}

In Table 2 we report summary statistics of the analyzed numerical variables.

\begin{table}[!htbp] \centering 
	\label{} 
	\tiny
	\resizebox{\columnwidth}{!}{%
		\begin{tabular}{ c c c c c } \\  [-2.5ex]
			\textbf{Variable}& \textbf{Mean}&\textbf{Variance}&\textbf{Standard Deviation}&\textbf{Kurtosis}\\ \hline
			Capital to asset & 2.5 & 10.2 & 3.2 & 21.5\\
			Reserves to asset & 4.2 & 8.5 & 2.9 & 4.3\\
			Interest to liab & 3.4 & 8.5& 2.9 & 104.6\\
			Financial assets to gdp & 385.0 & 134,365.2& 366.6& 33.2\\
			Mortgages to loans d4 & 0.2 & 1.7 & 1.3 & 0.1\\
			Securities to liab d4 & -12.0 & 1,342,234.9 & 1,158.5 & 105.7\\
			Credit to gdp & 140.2 & 2,623.4 & 51.2 & 0.0\\
			Credit to gdp d12 & 13.7 & 479.1 & 21.9& 0.5\\
			House Price Index rt16 gap & -2.5 & 33.7& 5.8 & 6.8\\
			International Investment Position  & -21.0 & 2,967.7 & 54.5 & 0.0\\
			Private Debt& 188.2 & 4938.7 & 70.3 & 0.1\\
			Gov Bold Yield d4 & 0.0 & 11.4 & 3.4 & 23.5\\ \hline 
		\end{tabular} 
	}
	\caption{Summary statistics of available numerical variables. } 
\end{table}

Then, we match the distress events with the available textual news data. The events are based upon bankruptcies and direct defaults, government aid and distressed mergers as presented in (Betz et al. 2014).
The distress events in this dataset are of three types. The first type of events include bankruptcies, liquidations and defaults, with the aim of capturing direct bank failures. The second type of events comprises the use of state support to identify banks in distress. The third type of events consists of forced mergers, which capture private sector solutions to bank distress. The inclusion of state interventions and forced mergers is important to better represent bank distress since
there have been few European direct bank failures in the considered period. Bankruptcies occur if a bank net worth falls below the country-specific guidelines, whereas liquidations occur if a bank is sold and the shareholders do not receive full payment for their ownership. Defaults occur if a bank failed to pay interest or principal on at least one financial obligation beyond any grace period specified by the terms or if a bank completes a distressed exchange. The distress events are formally considered to start when a failure is announced and end at the time of the 'de facto' failure.

A capital injection by the state or participation in asset relief programs (i.e., asset protection or asset
guarantees) is an indication of bank distress. From this 'indicator' are excluded liquidity support and
guarantees on banks’ liabilities since they are not used for defining distressed banks. The starting dates
of the events refer to the announcement of the state aid and the end date to the execution of the state
support program.
Distressed mergers are defined to occur if (i) a parent receives state aid within 12 months after a
merger or (ii) if a merged entity exhibits a negative coverage ratio within 12 months before the merger.
The dates for these two types of distress events are defined as follows, respectively: (i) the starting date
is when the merger occurs and the end date when the parent bank receives state aid, and (ii) the start
date is when the coverage ratio falls below 0 (within 12 months before the merger) and the end
date when the merger occurs. Thus far, data at hand assign a unique label for the stress events, not allowing 
a more detailed descriptive summary of the three event types.

\subsection{Data Integration}

The following step in the data preparation has been the addition of numerical financial data to the text news database. The numerical data were aligned with the set of sentences in which bank names occurred. The purpose was to match each and every mention of a bank with the corresponding numerical financial data aligned according to the same time horizon. Since the news and the financial data have different frequency, in particular, news have higher frequency while financial data are reported quarterly, the latter are replicated several times to perfectly match with the former. For each news regarding a bank within a given quarter, financial data are replicated and appended to the semantic vector of the news. Such matching activity between numerical and textual data resulted in the removal of some banks from the dataset due to missing data, causing a reduction from 101 to 62 target institutes (Table 3) and from about 601k to 380k news sentences. After cleaning the dataset, numerical data have been normalized to improve the classifier training time, due to easier convergence of the model. We have normalized the data with a standard approach, that is, by subtracting the mean value from each numerical variable of the dataset and dividing it by the standard deviation.
The resulting input vector for the 612 dimensional input layer of the neural classifier is thus composed of a numerical vector obtained joining together the 600-dimensional semantic vector coming from the unsupervised modelling, described in section 2, with the 12-dimensional numerical financial data vector. 
The dataset is then split into five folds, three for training, one for validation and one for testing according to a cross-validation scheme. The folds are created so that all the data regarding a given bank are in the same fold. 
The framework we apply is composed of an unsupervised algorithm and a supervised neural network classifier. To train the classifier a label indicating the distressed or tranquil status of the bank is provided. The dataset has been labelled according to the bank status with 0 for tranquil and 1 for distress. The proportions of the two classes are highly unbalanced: $93\%$ of the data-points represent tranquil status and only the remaining $7\%$ associated to distress events. Such imbalance of the classes has a significant impact both on the training and on the evaluation of the model. Regarding the training, it is important that the model is able to generalize due to the few distress examples, while for the evaluation it is important to provide an alternative measure to the accuracy. Using such index, in fact, a trivial model that always predicts the tranquil status would achieve a $93\%$ accuracy. Thus, it is necessary to measure an improvement against this baseline. Moreover the user is likely interested in weighting differently first error and second error types, especially in early warning applications. The usefulness measure, introduced in (Sarlin 2013), is able to handle these requirements.
\\

\begin{table}[!htbp] \centering 
	
	\label{} 
	\tiny
	\resizebox{\columnwidth}{!}{%
		\begin{tabular}{ c c c c c c } 
			\\[-2.8ex]
			\textbf{Financial Institution}& \textbf{Country} & \textbf{Financial Institution}& \textbf{Country} & \textbf{Financial Institution}& \textbf{Country}  \\ 
			\hline \\[-1.8ex] 
			Aareal Bank & DE & Carnegie Investment Bank & SE & Kommunalkredit& AT  \\ 
			ABN Amro & NL  & Commerzbank  & DE & LBBW & DE  \\ 
			Agricultural Bank of Greece & GR & Credit Mutuel & FR & Lloyds TSB & UK \\ 
			Allied Irish Banks & IE & Credito Valtellinese & IT  & Max Bank & DK  \\ 
			Alpha Bank & GR  & Cyprus Popular & CY  & Monte dei Paschi di Siena & IT    \\ 
			Amagerbanken & DK  & Danske Bank & DK  & National Bank of Greece & GR \\ 
			ATE Bank & GR & Dexia & FR & Nordea & SE  \\ 
			Attica Bank & GR  & EBH & DK  & NordLB & DE  \\ 
			Banca Popolare di Milano & IT  & EFG Eurobank & GR  & Nova ljubljanska banka Group (NLB) & SI \\ 
			Banco Popolare & IT  & Erste Bank & HU   & OTP Bank Nyrt & HU \\ 
			Bank of Cyprus Public Co Ltd & CY  & Fionia (Nova Bank) & DK  & Piraeus Bank & GR, CY  \\ 
			Bank of Ireland & IE   & Fortis Bank & LU, NL, BE  & Pronton Bank & GR  \\ 
			Banque Populaire & FR   & HBOS & UK  & RBS & UK  \\ 
			Bawag & AT   & Hellenic & GR  & Roskilde Bank & DK \\ 
			BayernLB & DE   & HSH Nordbank & DE  & Societe Generale & FR   \\ 
			BBK & ES   & Hypo Real Estate & DE  & Swedbank & SE   \\
			BNP Paribas & FR  & Hypo Tirol Bank & AT  & T-Bank & GR  \\  
			BPCE & FR & IKB & DE  & UNNIM & ES     \\ 
			Caixa General de Depositos & PT   & ING & NL   & Vestjysk & DK   \\ 
			Caja Castilla-La Mancha & ES  & Irish Nationwide Building Society & IE  &  &     \\ 
			CAM & ES  & KBC & BE  & &     \\ \hline 
		\end{tabular} 
	}
	\caption{List of considered financial institutions }
\end{table}

\section{Results}

The experimental results confirm that the integration of numerical and textual data amplifies the prediction capability of the model compared to the inclusion of only textual data. The distress events in the database represent only 7\% of the cases, resulting in very skewed training classes as explained earlier. Moreover, given the nature of the problem, the identification of distress situations, it could be useful to weight differently false positives and false negatives. In an early warning application, a sensitive system is often preferable since a further investigation phase follows the detection of possible events. These peculiarities have to be taken into account during the evaluation of the model.

\subsection{Evaluation and experimental results}
To manage the peculiarities of the evaluation of our particular problem we resort to the relative usefulness as measure of performance.
The relative usefulness ($U_r$), introduced in (Sarlin 2013) is a measure that allows to set the error type preference ($\mu$) and to measure the relative performance gain of the model over the baseline compared to a perfect model. 
Such index is based on the appropriate combination of the probabilities of the true positive ($TP$), false positive ($FP$), true negative ($TN$) and  false negative ($FN$) that generate the model loss $L_m$  (eq. 4) and on a  baseline loss $L_b$ set to be the best guess according to prior probabilities $p(obs)$ and error preferences $\mu$ (eq. 3). 

\medskip

\begin{equation}
L_b = min
\bigg \{
\begin{array}{l}
\mu * p(obs=1)\\
(1-\mu)* p(obs=0) \\
\end{array}
\end{equation}

\medskip

\begin{equation}
L_m= \mu * p(FN)+ (1-\mu)* p(FP)
\end{equation}

The absolute Usefulness ($U_a$) and the relative Usefulness ($U_r$) are directly derived from the loss functions: 

\begin{equation}
U_r=\frac{U_a}{L_b}=\frac{L_b-L_m}{L_b}
\end{equation}

As we can see from eq. 5 the relative usefulness is equal to 1 when the model loss ($L_m$) is equal to 0, thus the model is perfect. As a consequence, the relative usefulness measures the gain over the baseline compared to the gain that an ideal model would achieve. To compute the relative usefulness ($U_r$) we have set the error type preference ($\mu$) equal to 0.9 in accordance with the indications of previous studies like (Betz et al. 2014) and (Constantin et al. 2017) on the importance of signalling every possible crisis at cost of some false positive ($FP$) (with $\mu$ = 0.9 we are saying that missing a crisis is about 9 times worse than falsely signalling one). This is especially true if following the signalling of an event, a further investigation action is triggered.
In order to evaluate distress condition of a bank over a period, the predictions are aggregated on a monthly basis by bank entity. This is done averaging the predictions at the single sentence level by month for each different bank. This has been done to take into account the information available over one month period reducing the predictions variability. As a result of this procedure, the classification task can be summarized as understanding which banks are in distress status month by month based on the news and numerical data available over the previous month.
\\
\\
To evaluate the model on this classification task, we have trained it fifty times on the same dataset, recording the relative usefulness ($U_r$) result after each run and then averaging them. For each of the fifty trainings, the folds are resampled and the neural net is randomly initialized. To quantify the gain obtained from merging numerical and textual data we have done three different experiments, running the model respectively with textual data only (fig. 2, left), numerical data only (fig. 2, center) and numerical and textual data together (fig. 2, right).
As it is possible to see in fig. 2 the case with textual data alone achieves an average relative usefulness of 13.0\%, while the case with numerical data alone shows an average relative usefulness of 31.1\%. The combination of these two dataset and their exploitation in the model grants an average relative usefulness of 43.2\%, thus it positively enhances the prediction capability of the model. From these results we can also understand that, as expected, the financial numerical data hold the majority of the informative potential necessary for the labelling task but that the addition of textual information provides a non-negligible 12.1\% improvement to the relative usefulness of the model. 

\begin{figure}
\centering
\includegraphics[scale=0.5]{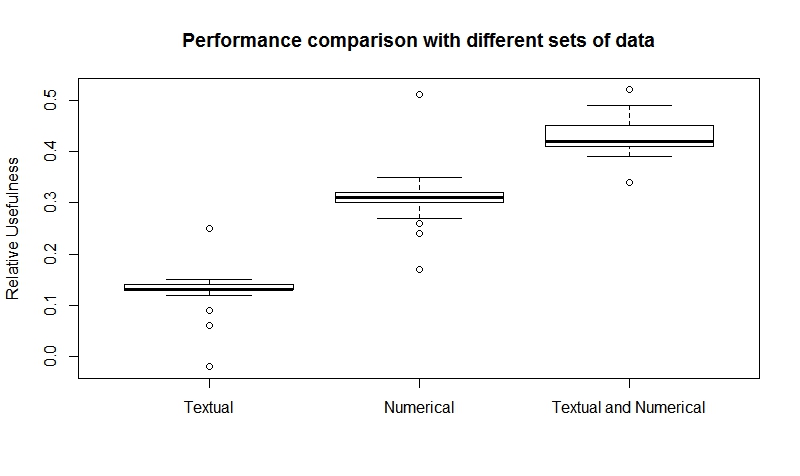}
\caption{Comparison of the relative usefulness obtained with the textual financial data (left), numerical financial dataset (center) and with their combination (right)}
\end{figure} 

\subsection{Classifier tuning}
In order to improve the classifier performance we have run a sensitivity analysis exploring different neural network configurations while training it with the Nesterov Accelerated Gradient Descent algorithm from (Nesterov 1983). We have tested different hidden layers dimensionalities, numbers of layers, learning rates, regularization parameters and dropout fractions (Hinton et al. 2012). For choosing the network configuration, we apply the Occam’s razor principle always preferring the simpler structure able to achieve a given performance. Thus, where performance is not reduced excessively, we try to select the network structure with fewer layers and fewer hidden nodes; this also helps to have better generalizing model and avoid overfitting when applying it to other datasets. In terms of hidden layers number, the network with one hidden layer (three layers in total including input and output) performs slightly better than those with more layers. We tested up to three hidden layers (5 layers in total) and the performance was monotonically decreasing. Regarding the number of hidden nodes, the network that gave the best usefulness has 50 hidden nodes with a learning rate $\alpha$ of 5$e$-4 combined with an $L_1$ regularization parameter $\lambda$ of 1$e$-5. The parameter that mostly affects the results is the number of nodes in the hidden layer. The results of the sensitivity analysis on the hidden nodes number (with regards to one hidden layer network configuration) are reported in fig.3, 4 and 5 respectively for the case including textual data alone, financial numerical data alone and the combination of the two. The range of hidden nodes in the three sensitivities is different because the input vectors in the three cases have very different dimensionalities, 600 input nodes when considering only textual data, 12 input nodes when considering only numerical data and 612 input nodes when including both numerical and textual data. We do not investigate extensively the textual data case which has already been studied in (R\"onnqvist and Sarlin 2017). Regarding the numerical based case, we can notice that we have a range of hidden layer size comprised between 10 and 20 nodes where performances are stable and Relative Usefulness is around 30\%. For the combined dataset (Numerical and Textual) we can see that there is a range around 50-60 hidden nodes where performance is stable around a 40\% Relative Usefulness. We expected the right number of hidden nodes to be similar to the Textual data case since the input dimensionality is similar (600 and 612).

\begin{figure}[!htbp]
\centering
\includegraphics[scale=0.5]{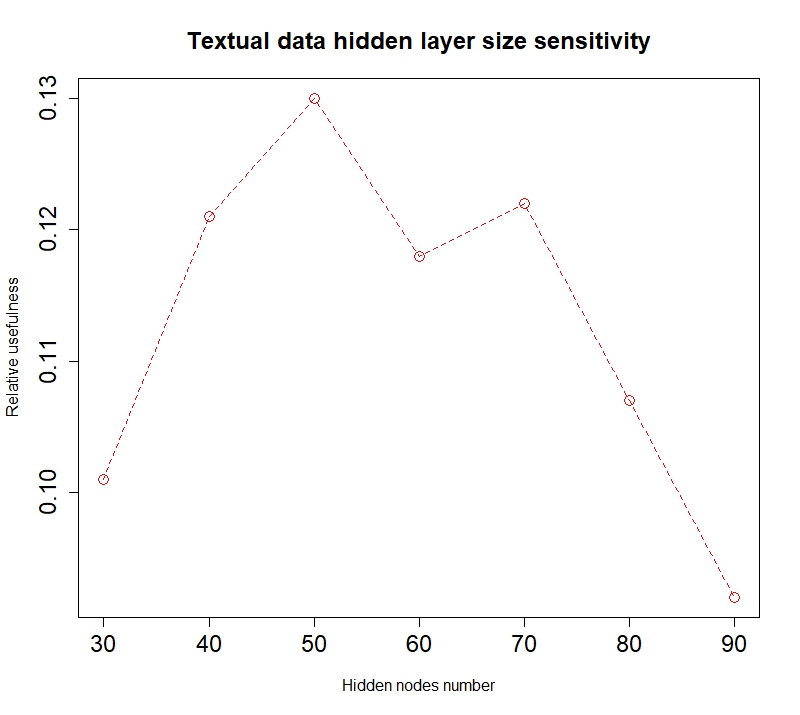}
\caption{Textual data - sensitivity analysis on the number of nodes of the hidden layer}
\end{figure}

\begin{figure}[!htbp]
\centering
\includegraphics[scale=0.5]{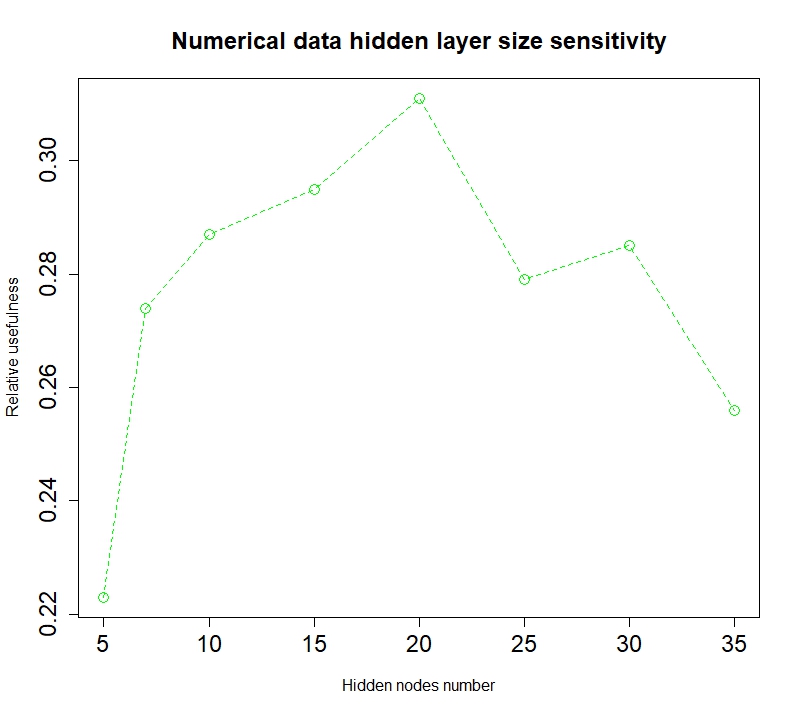}
\caption{Numerical data - sensitivity analysis on the number of nodes of the hidden layer}
\end{figure}

\begin{figure}[!htbp]
\centering
\includegraphics[scale=0.5]{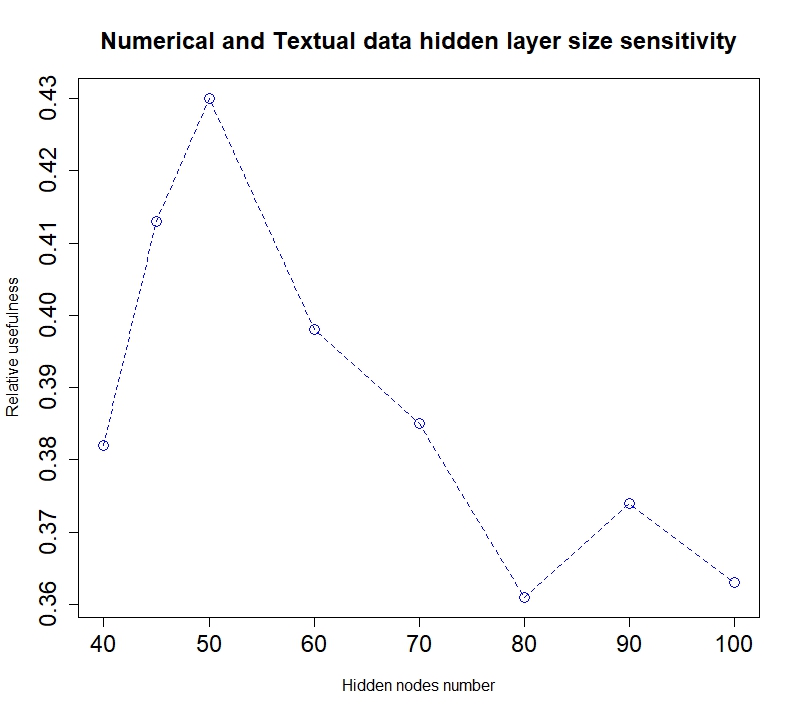}
\caption{Numerical and Textual data - sensitivity analysis on the number of nodes of the hidden layer}
\end{figure}

\section{Concluding Remarks}
\label{sec:Conclusion} 

In this work we have presented an approach for the integration of financial numerical data and financial news textual data into a single machine learning framework. The aim is to identify banks' distress conditions with improved performance with respect to the model based on news data only. The implemented framework processes textual data through an unsupervised neural network model converting the documents sentences into sentence vectors. The retrieved sentence vectors can then be joined with the financial numerical data in a unique input vector and fed to a supervised classifier, in our case a three layers fully connected neural network.
\\
The classification task was characterized by a high imbalance in the classes which poses concerns for both the training and the evaluation of the model. However, the implemented neural network is able to learn the combinations of the financial conditions of the banks and the semantic content of the news that are more associated with distress conditions. This is reflected in the improved performance obtained when leveraging both textual news data and financial numerical data, average relative usefulness of 43.2\%, compared to 31.1\% when using only numerical data and 13.0\% when utilizing only textual data.

Some limitations of this model reside in the way the news are processed and converted into vectors and how they are fed to the network to classify the distress. Methods like Doc2Vec with Distributed Memory approach in fact, while not using a pure bag of word approach, still ignore important text information to truly understand a sentence and not only its topic or its average sentiment. For example exact sequence patterns or long range dependencies in the text are not considered. Moreover Doc2Vec performs significantly better when trained on a large quantity of text similar to the application domain. This quantity of texts was available in our study but could pose a limit to applications in niche specific domains or its extension to less widespread languages. 

Anyway, in the last years there have been many improvements in the NLP field that can help overcome these limitations. A particularly interesting tool is the so-called Sequence to Sequence RNN class of models, that recently has become very popular. These models are composed of two RNNs (one encoder and a decoder) that are trained in an unsupervised setting to reconstruct their own input text. Sequence to Sequence architectures pretrained on financial and bank related text could be used as a substitute for the Doc2Vec representation our approach. Differently from Doc2Vec this kind of models consider more extensively word order and longer range dependencies between words when computing the text vector representation. Moreover, it is possible to augment the model capability providing few additional manually engineered features like a gazette of words with positive/negative polarity from a financial stability point of view.

Another future work direction that could improve this model as an early warning tool would be considering the news dynamic evolution. In this work, news are aggregated at monthly level, thus sub-monthly dynamics are lost. Using a RNN as distress classifier, it would be possible to sequentially feed the news vectors into the network taking also these dynamic effects into account (e.g. overall negative sentiment but with a positive trend in the last weeks).}

The methodology here applied is general and extensible to other problems were the integration of text and numerical covariates can improve classification and early warning performances. Other interesting results are to be expected in areas where textual data hold information with higher granularity and frequency, directly influencing the data to be predicted in the short run like in the case of financial markets.

\section*{References}

Y. Bengio, R. Ducharme, P. Vincent, and C. Janvin. A neural probabilistic language model. The Journal of Machine Learning Research, 3:1137–1155, 2003.

F. Betz, S. Oprică, T. A. Peltonen, and P. Sarlin. Predicting distress in European banks. Journal of Banking \& Finance, 45:225–241, 2014.

D. Bholat, S. Hansen, P. Santos, and C. Schonhardt-Bailey. Text mining for central banks. In Centre for Central Banking Studies Handbook, volume 33. Bank of England, 2015.

D.M. Blei, A.Y. Ng, and M.I. Jordan. Latent Dirichlet allocation. JMLR, 3:993–1022, 2003.

P. F. Brown, V. J. D. Pietra, P. V. deSouza, J. C. Lai and R. L. Mercer. Class-based n-gram models of natural language. Computational Linguistics, 18(4):467–479, 1992.

P. Cerchiello, P. Giudici and G. Nicola. 2017. Twitter data models for bank risk contagion, In Neurocomputing, Volume 264, 2017, Pages 50-56, ISSN 0925-2312, https://doi.org/10.1016/j.neucom.2016.10.101.

K. Cho, B. van Merrienboer, C. Gulcehre, F. Bougares, H. Schwenk, and Y. Bengio. Learning phrase representations using RNN encoder-decoder for statistical machine translation. In EMNLP, 2014.

A. Clark. Combining distributional and morphological information for part of speech induction. In Proc. of EACL, 2003.
S. Hochreiter, J. Schmidhuber. Long Short-Term Memory. Neural Computation 9(8):1735-1780, 1997

R. Collobert, J. Weston, L. Bottou, M. Karlen, K. Kavukcuoglu and P. Kuksa. Natural Language Processing (Almost) from Scratch. Journal of Machine Learning Research, 12:2493- 2537, 2011.

A. Constantin, T. Peltonen, P. Sarlin. Network linkages to predict bank distress. Journal of Financial Stability, 2016.

I. Guyon, A. Elisseeff. An introduction to variable and feature selection. Journal of Machine Learning Research, 2003, vol. 3 (pg. 1157-1182)

G. E. Hinton, J. L. McClelland, and D. E. Rumelhart. Distributed representations. In Rumelhart, D. E. and McClelland, J. L., editors, Parallel Distributed Processing: Explorations in the Microstructure of Cognition. 1986. Volume 1: Foundations, MIT Press, Cambridge, MA. pp 77-109.

G. Hinton, N. Srivastava, A. Krizhevsky, I. Sutskever, R. Salakhutdinov. 2012. Improving neural networks by preventing co-adaptation of feature detectors. CoRR, abs/1207.0580.

R. Hodrick, E.C. Prescott. Post-war U.S. business cycles: An empirical investigation.
Mimeo. Carnegie-Mellon University, Pittsburgh, PA (1980)

J. Hokkanen, T. Jacobson, C. Skingsley, and M. Tibblin. The Riksbank’s future information supply in light of Big Data. In Economic Commentaries, volume 17. Sveriges Riksbank, 2015.

N. Kalchbrenner, E. Grefenstette, P. Blunsom. 2014. A Convolutional Neural Network for Modelling Sentences. In Proceedings of ACL 2014.

T. Landauer, P.W. Foltz, D. Laham. Introduction to Latent Semantic Analysis. Discourse Processes. 25 (2–3): 259–284, 1998. doi:10.1080/01638539809545028 

Q. Le, T. Mikolov. Distributed Representations of Sentences and Documents. Proceedings of the 31 st International Conference on Machine Learning, Beijing, China, 2014. JMLR: W\&CP, volume 32.

P. Malo, A. Sinha, P. Korhonen, J. Wallenius, and P. Takala. Good debt or bad debt: Detecting semantic orientations in economic texts. Journal of the Association for Information Science and Technology, 65(4):782–796, 2014.

S. Martin, J. Liermann and H. Ney. Algorithms for bigram and trigram word clustering. Speech Communication, 24, 19–37, 1998.

T. Mikolov. Statistical Language Models Based on Neural Networks. PhD thesis, PhD Thesis, Brno University of Technology, 2012.

T. Mikolov, K. Chen, G. Corrado, and J. Dean. Efficient estimation of word representations in vector space. In Proceedings of Workshop at International Conference on Learning Representations, 2013.

Y. Nesterov. A method of solving a convex programming problem with convergence rate o (1/k2). In Soviet Mathematics Doklady, volume 27, pages 372–376, 1983.

R. Nyman, D. Gregory, K. Kapadia, P. Ormerod, D. Tuckett, and R. Smith. News and narratives in financial systems: exploiting big data for systemic risk assessment. BoE, mimeo, 2015.

J. Pennington, R. Socher, and C. Manning. 2014. Glove: Global vectors for word representation. In Proceedings of the 2014 Conference on Empirical Methods in Natural Language Processing (EMNLP), pages 1532–1543, Doha, Qatar, October. Association for Computational Linguistics.

S. R\"onnqvist, P. Sarlin. 2017. Bank distress in the news: Describing events through deep learning, Neurocomputing, Volume 264, 2017, Pages 57-70, ISSN 0925-2312, https://doi.org/10.1016/j.neucom.2016.12.110.

P. Sarlin. On policymakers’ loss functions and the evaluation of early warning systems. Economics Letters, 119(1):1–7, 2013.

J. Schmidhuber. Deep learning in neural networks: An overview. Neural Networks, 61:85–117, 2015.

R. Socher, J. Pennington, E. H. Huang, A. Y. Ng, and Christopher D. Manning. 2011. Semi-Supervised Recursive Autoencoders for Predicting Sentiment Distributions. In Proceedings of the 2011 Conference on Empirical Methods in Natural Language Processing (EMNLP).

R. Socher, A. Perelygin, J. Wu, J. Chuang, C. D. Manning, A. Y. Ng, and Christopher Potts. 2013b. Recursive deep models for semantic compositionality over a sentiment treebank. In Proceedings of the 2013 Conference on Empirical Methods in Natural Language Processing, pages 1631–1642, Stroudsburg, PA, October. Association for Computational Linguistics.

C. K. Soo. Quantifying animal spirits: news media and sentiment in the housing market. Ross School of Business Paper No. 1200, 2013.

Conflicts of interest: none

\end{document}